\documentclass[runningheads,a4paper]{llncs}

\usepackage{amssymb}
\usepackage{graphicx}

\usepackage{url}

\usepackage{capt-of}
\usepackage{tabto}
\usepackage{siunitx}
\usepackage{textcomp}
\newcommand{\textapprox}{\raisebox{0.5ex}{\texttildelow}}
\graphicspath{ {./figures/} }

\begin{document}

\mainmatter  

\title{A Multi-Scale CNN and Curriculum Learning Strategy for Mammogram Classification}

\titlerunning{Multi-Scale CNN and Curriculum Learning for Mammogram Classification}

\author{William Lotter\inst{1,2}\and Greg Sorensen\inst{2}\and David Cox\inst{1,2}}
\index{Lotter, William}
\index{Sorensen, Greg}
\index{Cox, David}
\authorrunning{Lotter et al.}

\tocauthor{William Lotter, Greg Sorensen, David Cox}

\institute{Harvard University, Cambridge MA, USA \and DeepHealth Inc., Cambridge MA, USA,
\\
\email{\{lotter,sorensen,davidcox\}@deephealth.io}}

\maketitle
\begin{abstract}
Screening mammography is an important front-line tool for the early detection of breast cancer, and some 39 million exams are conducted each year in the United States alone.
Here, we describe a multi-scale convolutional neural network (CNN) trained with a curriculum learning strategy that achieves high levels of accuracy in classifying mammograms.
Specifically, we first train CNN-based patch classifiers on segmentation masks of lesions in mammograms, and then use the learned features to initialize a scanning-based model that renders a decision on the whole image, trained end-to-end on outcome data.
We demonstrate that our approach effectively handles the ``needle in a haystack'' nature of full-image mammogram classification, achieving 0.92 AUROC on the DDSM dataset.
\end{abstract}

\section{Introduction}
\vspace{-0.05in}
Roughly one eighth of women in the United States will develop breast cancer during their lifetimes~\cite{breastcancerorg}.
Early intervention is critical --- five-year relative survival rates can be up to 3-4 times higher for cancers detected at an early stage versus at a later stage~\cite{seer}.
An important tool for early detection is screening mammography, which has been attributed with a significant reduction in breast cancer mortality~\cite{Berry_2005}.
However, the overall value of screening mammography is limited by several factors: reading mammograms is a tedious and error-prone process, and not all radiologists achieve uniformly high levels of accuracy~\cite{Elmore_2009,BCSC_2017}.
In particular, empirically high false positive rates in screening mammography can lead to significant unnecessary cost and patient stress \cite{Brewer_2007,Myers_2015}.
For these reasons, effective machine vision-based solutions for reading mammograms hold significant potential to improve patient outcomes.

Traditional computer-aided diagnosis (CAD) systems for mammography have typically relied on hand-engineered features~\cite{Nishikawa_2007}. 
With the recent success of deep learning in other fields, there have been several promising attempts to apply these techniques to mammography~\cite{Arevalo_2016,Mordang_2016,Carneiro_2015,Dhungel_2015,Dhungel_2014,geras_2017,Kooi_2017,Levy_2016,Yi_2017,Zhu_2016}.
Many of these approaches have been designed for specific tasks or subtasks of a full evaluation pipeline, for instance, mass segmentation~\cite{Zhu_2016_v2,Dhungel_2014} or region-of-interest (ROI) microcalcification classification~\cite{Mordang_2016}. 
Here, we address the full problem of binary cancer status classification: given an entire mammogram image, we seek to classify whether cancer is present~\cite{Carneiro_2015,Kooi_2017,Zhu_2016}. 
As recent efforts have shown~\cite{geras_2017}, creating an effective end-to-end differentiable model, the cornerstone of supervised deep learning, is challenging given the ``needle in a haystack'' nature of the problem.  
To address this challenge, we have developed a two-stage, curriculum learning-based~\cite{Bengio_2009} approach.
We first train patch-level CNN classifiers at multiple scales, which are then used as feature extractors in a sliding-window fashion to build an image-level model.
Initialized with the patch-trained weights, the image-level model can then effectively be trained end-to-end.
We demonstrate the efficacy of our approach on the largest public mammography database, the Digital Database for Screening Mammography (DDSM)~\cite{DDSM}.
Evaluated against the final pathology outcomes, we achieve an AUROC of 0.92.
 \vspace{-0.05in}

\begin{figure}[h]
  \begin{center}
    \includegraphics[width=0.9\textwidth]{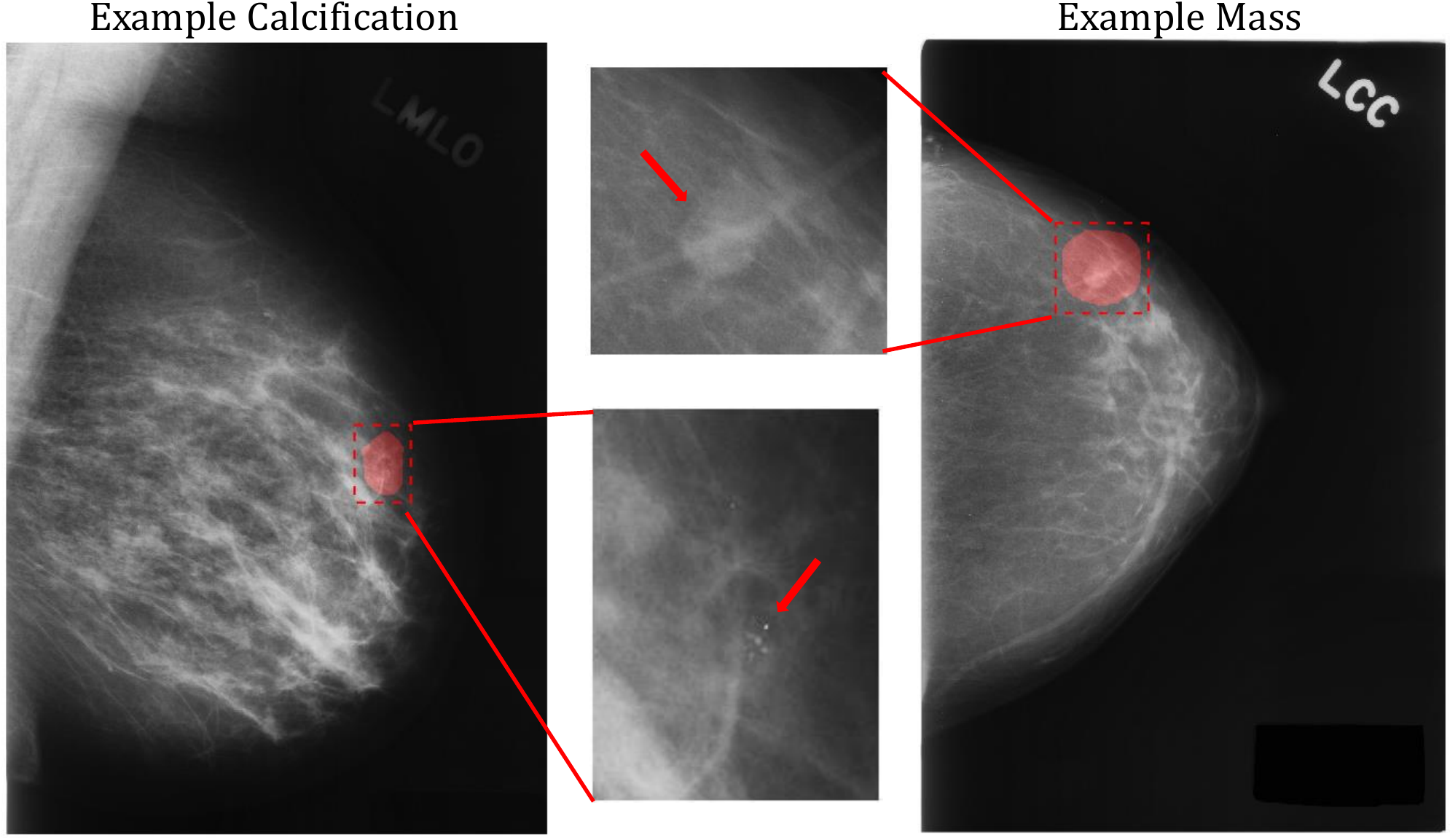}
  \end{center}
  \vspace{-0.22in}
  \caption{Examples of the two most common categories of lesions in mammograms, calcifications and masses, from the DDSM dataset~\cite{DDSM}. Radiologist-annotated segmentation masks are shown in red. The examples were chosen such that the mask sizes approximately match the median sizes in the dataset.}
  \label{examples}
\end{figure}
 \vspace{-0.25in}
 \vspace{-0.10in}
\section{Multi-Scale CNN with Curriculum Learning Strategy}
\vspace{-0.08in}
Figure~\ref{examples} shows typical examples of the two most common classes of lesions found in mammograms,  masses and calcifications.
Segmentation masks drawn by radiologists are shown in red ~\cite{DDSM}.
Even though the masks often encompass the surrounding tissue, the median size is only around $0.5\%$ of the entire image area for calcifications, and $1.2\%$ for masses.
The insets shown in Fig.~\ref{examples} illustrate the high level of fine detail required for detection.
As noted in~\cite{geras_2017}, the requirement to find small, subtle features in high resolution images (e.g. \textapprox5500x3000 pixels in the DDSM dataset) means that the standard practice of downsampling images to \textapprox250x250, which has proven effective in working with many standard natural image datasets~\cite{imagenet}, is unlikely to be successful for mammograms.  
It is for these reasons that traditional mammogram classification pipelines typically consist of a sequence of steps, such as candidate ROI proposals, followed by feature extraction, perhaps segmentation, and finally classification~\cite{Kooi_2017}.
While deep learning could in principle be used for any or all of these individual pieces, a general axiom of deep learning is that the greatest gains are seen when the system is trained ``end-to-end'' such that errors are backpropagated uninterrupted through the entire pipeline.

\vspace{-0.1in}
\begin{figure}[h]
  \begin{center}
    \includegraphics[width=1.0\textwidth]{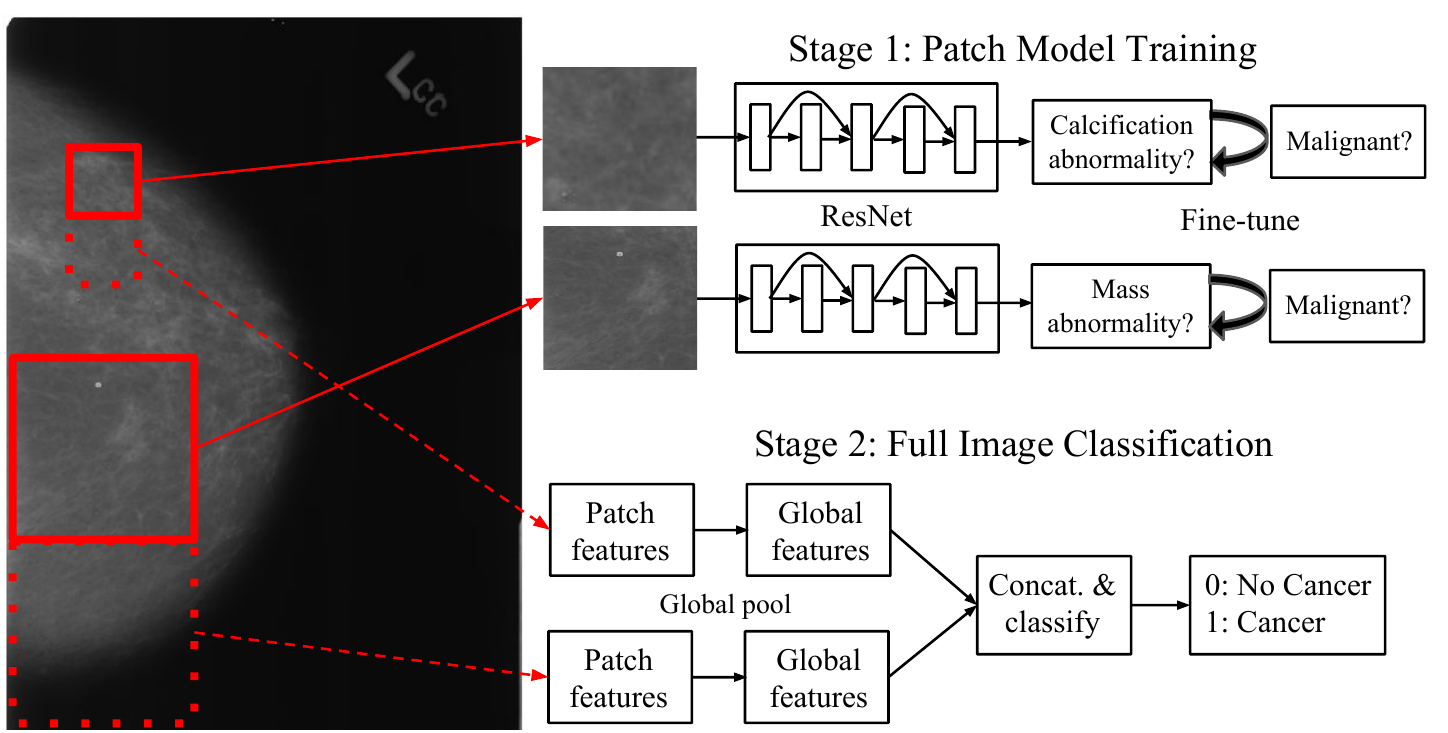}
  \end{center}
  \vspace{-0.24in}
  \caption{Schematic of our approach, which consists of first training a patch classifier, followed by image-level training using a scanning window scheme. We train separate patch classifiers for calcifications and masses, at different scales, using a form of a ResNet CNN~\cite{He_2015,Zagoruyko_2016}. For image-level training, we globally pool the last ResNet feature layer at each scale, followed by concatenation and classification, with end-to-end training on binary cancer labels.}
  \label{methods}
\end{figure}
\vspace{-0.11in}
 \vspace{-0.05in}
Our training strategy is illustrated in Fig~\ref{methods}.
The first stage consists of training a classifier to estimate the probability of the presence of a lesion in a given image patch. 
For training, we randomly sample patches from a set of training images, relying on (noisy) segmentation maps to create labels for each patch.
Given the different typical scales of calcifications and masses, we train a separate classifier for each.
We use ResNet CNNs~\cite{He_2015} for the classifiers, specifically using the ``Wide ResNet'' formulation~\cite{Zagoruyko_2016}.
We first train for abnormality detection (i.e. is there a lesion present), followed by fine-tuning for pathology-determined malignancy.

The second stage of our approach consists of image-level training.
Given the high level of scrutiny that is needed to detect lesions, and because of its compatibility with end-to-end training, we use a scanning-window (e.g. convolutional) scheme. 
We partition the image into a set of patches such that each patch is contained entirely within the image, and the image is completely tiled, but there is as minimal overlap and number of patches as possible.
Features are extracted for each patch using the last layer before classification of the patch model.
Final classification involves aggregation across patches and the two scales, for which we tried various pooling methods and number of fully-connected layers.
The strategy that ultimately performed best, as assessed on the validation data, was using global average pooling across patch features at each scale, followed by concatenation and a single softmax classification layer. 
Using a model of this form, we train end-to-end, including fine-tuning of the patch model weights, using binary image-level labels.

 \vspace{-0.08in}
\section{Experiments}
 \vspace{-0.08in}
We evaluate our approach on the original version of the Digital Database for Screening Mammography (DDSM)~\cite{DDSM}, which consists of $10480$ images from $2620$ cases.
Each case consists of the standard two views for each breast, craniocaudal (CC) and mediolateral-oblique (MLO).
As there is not a standard cross validation split, we split the data into an 87\% training/5\% validation/8\% testing split, where cross validation is done by patient.
The split percentages were chosen to maximize the amount of training data, while ensuring an acceptable confidence interval for the final test results.

For the first stage of training, we create a large dataset of image patches sampled from the training images.
We enforce that the majority of the patches come from the breast, by first segmenting using Otsu's method~\cite{Otsu}.
Before sampling, we resize the original images with different factors for calcification and mass patches.
Instead of resizing to a fixed size, which would cause distortions because the aspect ratio varies over the images in the dataset, or cropping, which could cause a loss of information, we resize such that the resulting image falls within a particular range.
We set the target size to 2750x1500 and 1100x600 pixels, for the calcification and mass scales, respectively.  
Given an input image, we calculate a range of allowable resize factors as the min and max resize factors over the two dimensions.
That is, given an example of size, say 3000x2000, the range of resize factors for the calcification scale would be [$1500/2000=0.75$ , $2750/3000=0.92$], from which we sample uniformly.
For other sources of data augmentation, we use horizontal flipping, rotation of up to $30^{\circ}$, and an additional rescaling by a factor chosen between 0.75 and 1.25.
We then sample patches of size 256x256.
In the first stage of patch classification training, lesion detection without malignancy classification, we create 800K patches for each lesion category, split equally between positive and negative samples.
In the second stage, we create 900K patches split equally between normal, benign, and malignant.

As mentioned above, we use ResNets~\cite{He_2015} for the patch classifiers.
We specifically use the ``Wide ResNet'' formulation~\cite{Zagoruyko_2016}, although, for the sake of training speed and to avoid overfitting, our networks are not particularly wide.
The Wide ResNet consists of groups of convolutional blocks with residual connections, and 2x2 average pooling between the groups. 
Each convolution in a block is proceeded by batch normalization~\cite{Ioffe_2015} followed by ReLU activation.
After the final group, features are globally average pooled, followed by a single classification layer.
The main hyperparameters of the model are the number of filters per layer and the number of residual blocks per group, $N$.
For our models, we use five groups with the number of filters per group of (16, 32, 48, 64, 96) and an $N$ of 2 and 4 for the calcification and mass models, respectively.
For more details of the architecture, the reader is referred to~\cite{Zagoruyko_2016}.
The only deviation we make is using 5x5 convolutions with a 3x3 stride for the initial convolutional layer, accounting for the relatively large input size we use of 256x256.

\begin{figure}[h]
  \begin{center}
    \includegraphics[width=0.98\textwidth]{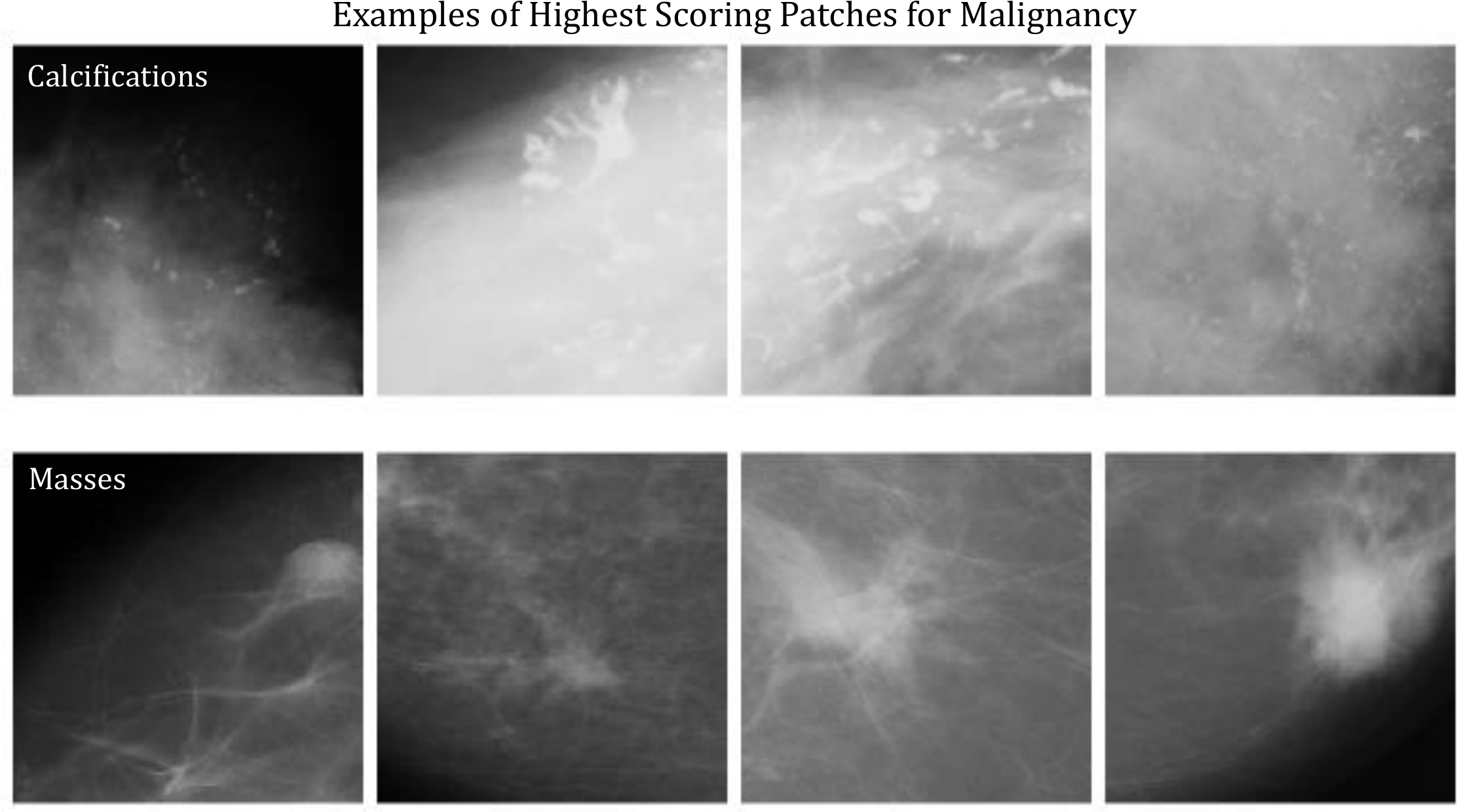}
  \end{center}
  \vspace{-0.25in}
  \caption{Examples of the highest scoring patches for malignancy in the test set. The top row corresponds to the calcification model and the bottom row corresponds to the mass model.}
  \label{patches_examples}
\end{figure}

For training the patch models, we use RMSProp~\cite{rmsprop} with a learning rate of \num{2e-4} and batch size of 32.
In the initial abnormality detection stage, we train for 50 epochs with 10K patch samples per epoch and an equal proportion of positive and negative samples, followed by 125 epochs with 15K per epoch and a positive sample rate of 25\%.
We then fine-tune for malignancy.
For the calcification model, we use a normal/benign/malignant labeling scheme.
We train for 225 epochs with 15K samples per epoch, and an equal proportion of the three classes.
The three-way labeling scheme caused overfitting with the mass model, so we instead use binary malignant/non-malignant labels.
The model is trained for 150 epochs with a sampling ratio of 20\% normal/40\% benign/40\% malignant.
To illustrate the information learned by the patch classifiers, we show several of the highest scoring patches for malignancy in the test set in Fig.~\ref{patches_examples}.

\vspace{-0.09in}
\vspace{-0.08in}

For the image-level training, we initialize the model with the final patch weights and follow a similar image resizing scheme.
We again augment using horizontal reflections and an additional rescaling by a factor chosen between 0.8 to 1.2.
At each scale, we divide the image into 256x256 patches, using the stride strategy explained earlier.
We also keep track of the regions of overlap between patches, and normalize these areas when global pooling, since otherwise the final features would be biased towards these locations.
For image-level labels, we categorize according to if there is a malignant lesion in either view of the breast.
Due to the possible different number of patches per image and because of the high memory footprint, we use a batch size of 1 during training.
We train for 100K iterations using RMSProp~\cite{rmsprop} with a learning rate of \num{2e-4}, followed by \num{4e-5} for 50K iterations.
Final weights were chosen by monitoring the area-under-the-curve (AUC) for a receiver operating characteristic (ROC) plot on the validation set.
While we train on a per image basis, we report final results on a (patient, laterality) basis by averaging final scores across the CC and MLO views of the breast.
For final test results, we average predictions across both horizontal orientations and five resize factors, chosen equally spaced between the allowable factors per image.

\begin{figure}[t]
	\centering
	\begin{minipage}[h]{0.6\textwidth}
		\centering
		
		\hspace*{-2.25in}
		\includegraphics[width=\textwidth]{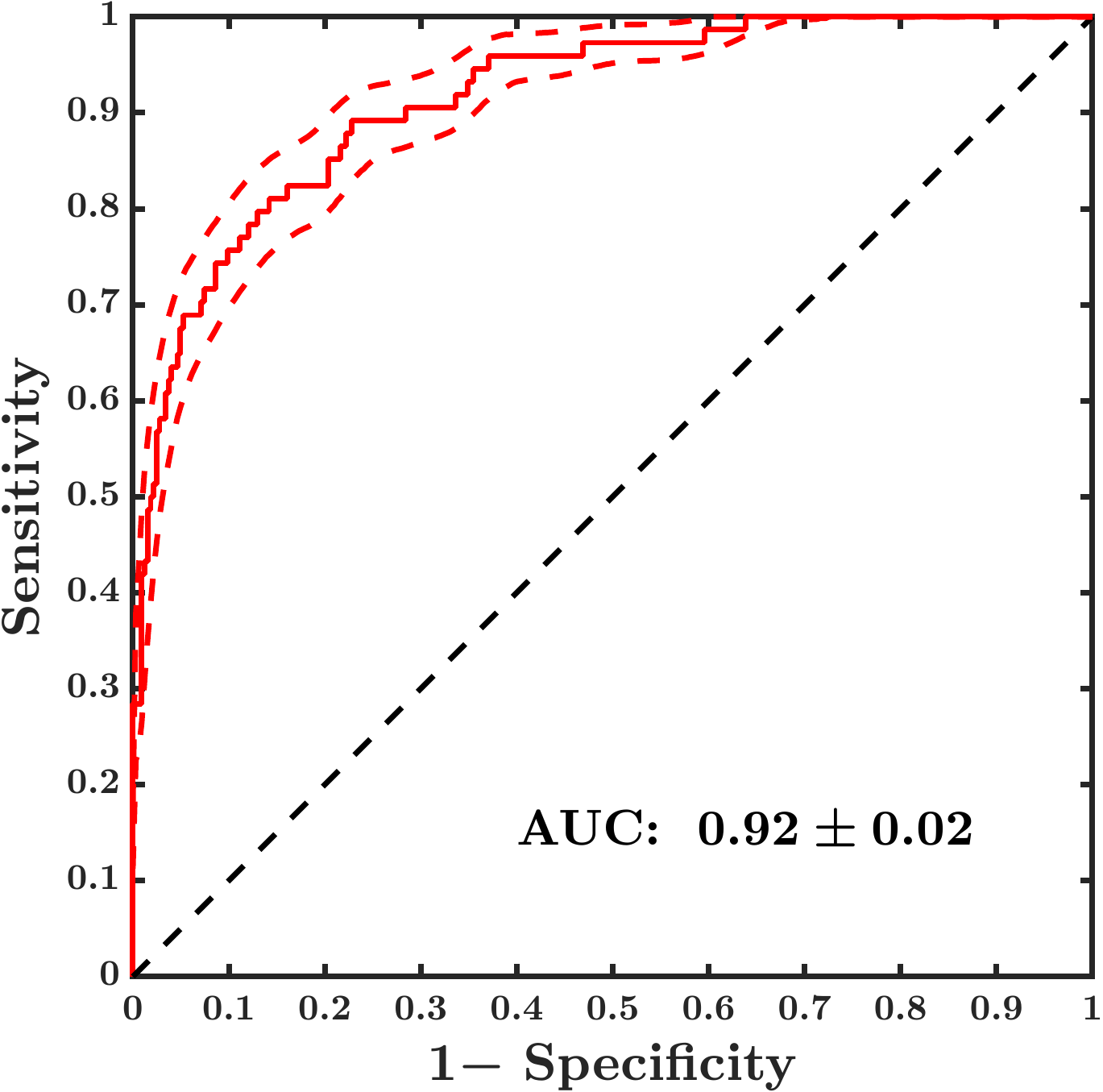}
		
	\end{minipage}\hfill
	\begin{minipage}[h]{0.37\textwidth}

	\hspace*{-1in}
	\parbox{1.8\textwidth}{\textbf{Fig. 4.} \textit{Left}: ROC curve for our model on a test partition of the DDSM dataset. Predictions and ground-truth are compared at a breast-level basis. \textit{Below}: ROC AUC by pre-training and data augmentation. InceptionV3 assumes a fixed input size, so ``size'' augmentation, i.e. random input image resizing, isn't directly feasible. ROC curve on left corresponds to the top row. \vspace*{0.1in}}
	
	\hspace*{-1in}
	\begin{tabular}{lccc}
                                                            & Pre-Training                                                    & \begin{tabular}[c]{@{}c@{}}Data\\ Augment.\end{tabular}        & AUC \\ \hline
\multicolumn{1}{l|}{Multi-scale CNN   \hspace{1pt}} & \multicolumn{1}{c|}{\hspace{1pt} DDSM lesions \hspace{1pt}} & \multicolumn{1}{c|}{\hspace{1pt} size, flips \hspace{1pt}} & \hspace{1pt} 0.92 $\pm$ 0.02                                                 \hspace{1pt}                    \\
\multicolumn{1}{l|}{Multi-scale CNN }                 & \multicolumn{1}{c|}{DDSM lesions}                               & \multicolumn{1}{c|}{flips}                                     & 0.89 $\pm$ 0.02                                                  \\
\multicolumn{1}{l|}{Multi-scale CNN }                 & \multicolumn{1}{c|}{none}                                       & \multicolumn{1}{c|}{flips}                                     & 0.65 $\pm$ 0.04                                                  \\
\multicolumn{1}{l|}{InceptionV3}                            & \multicolumn{1}{c|}{ImageNet}                                   & \multicolumn{1}{c|}{flips}                                     & 0.77 $\pm$ 0.03                                                  \\
\multicolumn{1}{l|}{InceptionV3}                            & \multicolumn{1}{c|}{none}                                       & \multicolumn{1}{c|}{flips}                                     & 0.59 $\pm$ 0.04                                                  \\ \hline

\end{tabular}

	\end{minipage}
	
\end{figure}

Fig. 4 contains the ROC curve on the test set for our proposed model.
We obtain standard deviation error bars using a bootstrapping estimate.
Our model achieves an AUC of 0.92 $\pm$ 0.02.
As full image mammogram classification lacks a standardized evaluation framework, it is somewhat difficult to directly compare our results to other work.
Related papers include the work of Carneiro et al.~\cite{Carneiro_2015}, Zhu et al.~\cite{Zhu_2016}, Kooi et al.~\cite{Kooi_2017}, and Geras et al.~\cite{geras_2017}.
Carneiro et al. use the InBreast~\cite{inbreast} dataset and a different version of DDSM, and use radiologist segmentation masks as input into their final model.
Zhu et al. use InBreast for mass classification, and do not rely on segmentation masks for training or inference.
Kooi et al. and Geras et al. both use private datasets.  

To provide some form of a meaningful comparison to our model, we report results here using a CNN designed for ImageNet classification.
We use the InceptionV3 model~\cite{Szegedy_2015,Szegedy_2015_v2}, choosing this model over alternatives because its input size is relatively large at 299x299.
Because InceptionV3 is designed for a fixed input size, training with resizing augmentation isn't feasible, but we do train with horizontal flip augmentation.
Consistent with many other results in the literature~\cite{Zhu_2016,Carneiro_2015}, we find that ImageNet pre-training of InceptionV3 helps for eventual training on the DDSM dataset.
However, the InceptionV3 model still underperforms our model, achieving an AUC of 0.77 $\pm$ 0.03. 
Without ImageNet pre-training, the InceptionV3 model achieves an AUC of 0.59 $\pm$ 0.04.
In both cases, results are still reported on a (patient, laterality) basis with averaging across views and possible horizontal orientations.
A summary of the results is contained in the table by Fig. 4.
To make a more controlled comparison, we also report the results for our model without size augmentation training. 
The performance drops slightly to 0.89 $\pm$ 0.02, but is still significantly higher than the InceptionV3 model.
The third row of the table also contains results for our model without the DDSM lesion pre-training, which decreases performance to 0.65 $\pm$ 0.04, however the model still performs slightly better than the InceptionV3 model without pre-training (last row).
Altogether, these results suggest that all elements of our approach --- including model formulation and pre-training scheme --- are important for accurate full image mammogram classification performance.

 \vspace{-0.07in}
\section{Conclusions}
 \vspace{-0.07in}
Computer-aided diagnosis for mammography is a heavily studied problem given its potential for large real-world impact.
This field, like many others, is transitioning from hand-engineered features to features learned in a deep learning framework.
While there have been many efforts to apply deep learning to subcomponents of the mammography pipeline, here we are concerned with full image classification.
Given the high resolution and relatively small ROIs, effectively designing an end-to-end solution is challenging.
We have presented a multi-scale CNN scanning window scheme with a lesion-specific curriculum learning strategy that achieves promising results.
Our approach performs significantly better than standard ``out of the box'' CNN models, and our experiments show that both the choice of architecture and training scheme play an important role in achieving this performance.
Future work includes learning interest points in a more unsupervised way, to reduce reliance on hand-drawn segmentation masks.
Overall, we argue that mammogram classification is a task that is well matched to the capabilities of modern deep learning approaches, and that it can serve as a natural testbed for the development of new deep learning techniques in the context of an application with clear potential for societal impact.
\vspace{-0.07in}

\bibliographystyle{splncs}
 

\end{document}